\pdfoutput=1

\documentclass{article}

\usepackage[final]{nips_2018}

\usepackage[utf8]{inputenc} % allow utf-8 input
\usepackage[T1]{fontenc}    % use 8-bit T1 fonts
\usepackage{hyperref}       % hyperlinks
\usepackage{url}            % simple URL typesetting
\usepackage{booktabs}       % professional-quality tables
\usepackage{amsfonts}       % blackboard math symbols
\usepackage{nicefrac}       % compact symbols for 1/2, etc.
\usepackage{microtype}      % microtypography
\usepackage{graphicx}
\usepackage{amsmath,amssymb} % define this before the line numbering.
\usepackage{color}
\usepackage{soul}
\usepackage[dvipsnames]{xcolor}
\usepackage{float}
\usepackage{algorithm}
\usepackage[noend]{algpseudocode}
\usepackage{subcaption}
\usepackage{multirow}
\usepackage{todonotes}
\usepackage{setspace}
\usepackage{comment}
\setcitestyle{square}
\usepackage{xspace}

\usepackage{xcolor}
\hypersetup{
    colorlinks,
    linkcolor={red!50!black},
    citecolor={blue!50!black},
    urlcolor={blue!80!black}
}

\title{Multi-Fidelity Recursive Behavior Prediction} % for autonomous driving applications}

\author{
    Mihir Jain \\
    Qualcomm AI Research
    \And 
    Kyle Brown$^{*}$ \\
    Stanford University \\
    \And Ahmed K. Sadek \\
    Qualcomm AI Research \\
}

\newcommand{\carstate}{\mathbf{x}} % car state

\newcommand{\initialtime}{t_0}
\newcommand{\currenttime}{t}
\newcommand{\finaltime}{t_f}
\newcommand{\policy}{\mathbf{\pi}}

\newcommand{\numcars}{n}

\newcommand{\level}{k}

\newcommand{\trafficscene}{\mathcal{X}}

\newcommand{\assignlevel}{\textproc{AssignReasoningLevel}}
\newcommand{\assignpolicies}{\textproc{AssignPolicyModels}}

\newcommand{\ourmodel}{MFRBP\xspace}

\newcommand{\CSP}{\policy_\text{CSP}} % convolutional social pooling
\newcommand{\FCCSP}{\policy_\text{FC-CSP}} % Future-conditional CSP
\newcommand{\CV}{\policy_\text{CV}} % constant velocity

\newcommand{\modelA}{L1-RBP\xspace}
\newcommand{\modelB}{L1-MFRBP\xspace}

\begin{document}

\maketitle

\begin{abstract}
 Predicting the behavior of surrounding vehicles is a critical problem in automated driving. 
    We present a novel game theoretic behavior prediction model that achieves state of the art prediction accuracy by explicitly reasoning about possible future interaction between agents.
        We evaluate our approach on the NGSIM vehicle trajectory data set and demonstrate lower root mean square error than state-of-the-art methods.
        
\end{abstract}
\let\thefootnote\relax\footnotetext{$^*$ Work done during Kyle Brown's internship at Qualcomm AI Research. Kyle's reseach is also supported by the National Science Foundation.}

%%%%%%%%%%%%%%%%%%%%
\section{Introduction}\label{sec:Introduction}
Predicting the future motion of surrounding vehicles is a critical problem in autonomous driving research and development.
    One of the key difficulties associated with behavior prediction is \emph{interaction} between traffic participants.
        Existing models vary in the way they reason about interactive driver behavior.
    Some models ignore interaction completely, predicting the future behavior of a target vehicle based solely on that vehicle's previous motion \citep{Ammoun2009,Barth2008,Hillenbrand2006,Kaempchen2004IMMOT,Miller,Polychronopoulos2007}.
    Other models \emph{implicitly} reason about interaction by conditioning motion prediction on the local traffic scene (including the current state or motion history of other nearby vehicles) \citep{Altche2018,Deo,Deo2018MultiModalTP,Khosroshahi,Morton2017,Phillips2017}.

    Still other models reason \emph{explicitly} about interaction, addressing the prediction task from a game-theoretic perspective  \citep{Galceran2017,HongYoo,Sadigh2016,Schmerling2017,Yoo2012}.

 We present a novel game theoretic behavior prediction model that we call Multi-Fidelity Recursive Behavior Prediction (\ourmodel{}). \ourmodel{} achieves better prediction accuracy (as measured by root mean squared error) than previous state-of-the-art models by explicitly reasoning about possible future interaction between agents.
    The proposed algorithm employs a recursive trajectory prediction scheme inspired by the Level-$k$ \citep{Costa-Gomes2001} and Cognitive Hierarchy \citep{Camerer2004} recursive reasoning paradigms.

This paper gives a condensed overview of the general Multi-Fidelity Recursive Behavior Prediction algorithm.
    We also discuss several specific implementations of our model, all of which incorporate specific elements from Convolutional Social Pooling for Vehicle Trajectory Prediction as proposed by Deo and Trivedi \citep{Deo} in 2018.
    All experiments are conducted with the publicly available NGSIM data set.

 %%%%%%%%%%%%%%%%%%%%%%%%%%%%%%%%%%
 
 \section{Methods}\label{sec:Methods}
Consider a traffic scene consisting of $\numcars$ agents.
    The motion history of the traffic scene from the initial time $\initialtime$ to the current time $\currenttime$ can be compactly represented by the set of time histories $\trafficscene_{\text{hist}} = \lbrace \carstate_1^{\initialtime:\currenttime}, \ldots,\carstate_\numcars^{\initialtime:\currenttime} \rbrace$, where $\carstate_i^{(\currenttime)}$ is the state of agent $i$ at time $\currenttime$.
        The corresponding set of future trajectories (from time $\currenttime+1$ to time $\finaltime$) is  $\trafficscene_{\text{future}} = \lbrace \carstate_1^{\currenttime+1:\finaltime}, \ldots, \carstate_\numcars^{\currenttime+1:\finaltime} \rbrace$.
            The input to our model is $\trafficscene_{\text{hist}}$, and the output is $\hat{P}(\trafficscene_{\text{future}})$, a probability distribution over the future trajectories of all agents.
                In all experiments presented here, we choose to model $\hat{P}(\trafficscene_{\text{future}})$ as a set of Gaussian trajectory predictions $\lbrace (\hat{\carstate}_{1}^{(\currenttime+1:\finaltime)}, \Sigma_{1}), \ldots,(\hat{\carstate}_{\numcars}^{(\currenttime+1:\finaltime)},\Sigma_{\numcars}) \rbrace$, where $\hat{\carstate}_i$ and $\Sigma_i$ represent the mean vector and covariance matrix of a Gaussian distribution over the future trajectory of the $i$th agent.
            
    The general Multi-Fidelity Recursive Behavior Prediction algorithm is outlined in Algorithm \ref{alg:general_MFRBP}.
        In the remainder of this section, we discuss two important characteristics of our model: \emph{recursive prediction} and \emph{multi-fidelity modeling}.

\vspace{-2mm}
\paragraph{Recursive Prediction}
Our model employs a game-theoretic recursive prediction scheme inspired by Level-$k$ Reasoning \citep{Costa-Gomes2001} and Cognitive Hierarchy \citep{Camerer2004}. Within this paradigm, a ``level $k$'' agent model assumes that all other agents act according to a level $k-1$ (or lower) agent model.

First, the algorithm assigns a reasoning level $\level_i$ and a corresponding sequence of policy models $(\pi_{i,0}, \ldots, \pi_{i,\level_i})$ to each agent $i \in \lbrace 1, \ldots, \numcars \rbrace$ in the scene.
    At each level $\level$ (starting from $\level = 0$ and proceeding upward) a predicted trajectory is generated for each agent $i \in \lbrace 1, \ldots, \numcars \rbrace$ if that agent's assigned reasoning level $\level_i$ is greater than or equal to $\level$.
        The crucial detail is that, for $\level>0$, the level $\level$ trajectory prediction for a given agent is \emph{explicitly conditioned} on the highest level (up to $\level-1$) previously computed trajectory predictions associated with each other agent.

    When the highest reasoning level has been reached (i.e. when $\level = \max_{i \in \lbrace 1, \ldots, \numcars \rbrace } \level_i$), the algorithm returns a set containing the final predicted trajectory for each agent.

\vspace{-2mm}
\paragraph{Multi-Fidelity Behavior Modeling}
For a given traffic scene history $\trafficscene_{\text{hist}}$, the output of our model depends entirely on the reasoning levels and sequences of policy models assigned to the agents.
    These assignments are determined by the user-defined methods on lines \ref{line:assign_level} and \ref{line:assign_policies} of Algorithm \ref{alg:general_MFRBP}.
        
In our experiments, we show that this design flexibility can be used for \emph{multi-fidelity modeling}, meaning higher-fidelity motion prediction for some agents than for others.
    \emph{Multi-fidelity modeling} can be useful in applications (e.g. automated driving) where we may know and/or care more about some agents than others.

\begin{algorithm}
    \caption{Multi-Fidelity Recursive Behavior Prediction}
    \label{alg:general_MFRBP}
    \begin{algorithmic}[1] % The number tells where the line numbering should start
        \Procedure{MultiFidelityRecursiveBehaviorPrediction}{$\trafficscene_{\text{hist}}$}
        \For{$i \in 1:\numcars$}
            \State $\level_i \gets \assignlevel(\trafficscene_{\text{hist}},i)$ \label{line:assign_level}
            \State $( \pi_{i,0}, \ldots, \pi_{i,\level_i} ) \gets \assignpolicies(\trafficscene_{\text{hist}},i,\level_i)$ \label{line:assign_policies}
            \State $(\hat{\carstate}_{i,0}^{(\currenttime+1:\finaltime)},\Sigma_{i,0}) = \policy_{i,0}
                        \big(
                            \lbrace
                            \carstate_{j}^{ (\initialtime:\currenttime)
                            } \mid j \in \lbrace 1, \ldots, \numcars \rbrace, j \neq i
                            \rbrace
                        \big)$ \label{line:level_0_policy}
        \EndFor
        \For{$k \gets 0, \ldots, \max_{i \in \lbrace 1, \dots, n \rbrace} \level_n$} \label{line:begin_recursion}
            \For{$i \in 1:\numcars$}
                \If{$k \leq \level_i$}
                    \State $(\hat{\carstate}_{i,k}^{(\currenttime+1:\finaltime)},\Sigma_{i,k}) = \policy_{i,k}\big( \lbrace 
                            (
                            \carstate_{j}^{
                            (\initialtime:\currenttime)},
                            \hat{\carstate}_{j,
                            min(\level_j,k-1)}^{
                            (\currenttime+1:\finaltime)}
                            )
                            \mid j \in \lbrace 1, \ldots, \numcars \rbrace, j \neq i
                        \rbrace \big)$ \label{line:level_k_policy}
                \EndIf
            \EndFor
        \EndFor \label{line:end_recursion}
        \State \textbf{return} $\lbrace (\hat{\carstate}_{1,\level_1}^{(\currenttime+1:\finaltime)}, \Sigma_{1,\level_1}), \ldots, 
        (\hat{\carstate}_{\numcars,\level_\numcars}^{(\currenttime+1:\finaltime)},
        \Sigma_{\numcars,\level_\numcars}) \rbrace$ \label{line:returnvalue}
        \EndProcedure
    \end{algorithmic}
\end{algorithm}

\vspace{-3mm}
\paragraph{Policy models used in our experiments} 
Our simple experiments incorporate three distinct policy models.
    The first two policy models condition only on motion \emph{history} $\trafficscene_{\text{hist}}$.
    whereas, the third policy model explicitly conditions on previously computed level $0$ trajectory predictions.

The \emph{Constant Velocity ($\CV$) model}
simply predicts that a target vehicle will travel at constant velocity equal to the average velocity vector (both longitudinal and lateral components) over the last second.
    This can be thought of as ``low-fidelity'' motion prediction.

\begin{figure}[!t]
    \centering
    \includegraphics[width=0.9\textwidth]{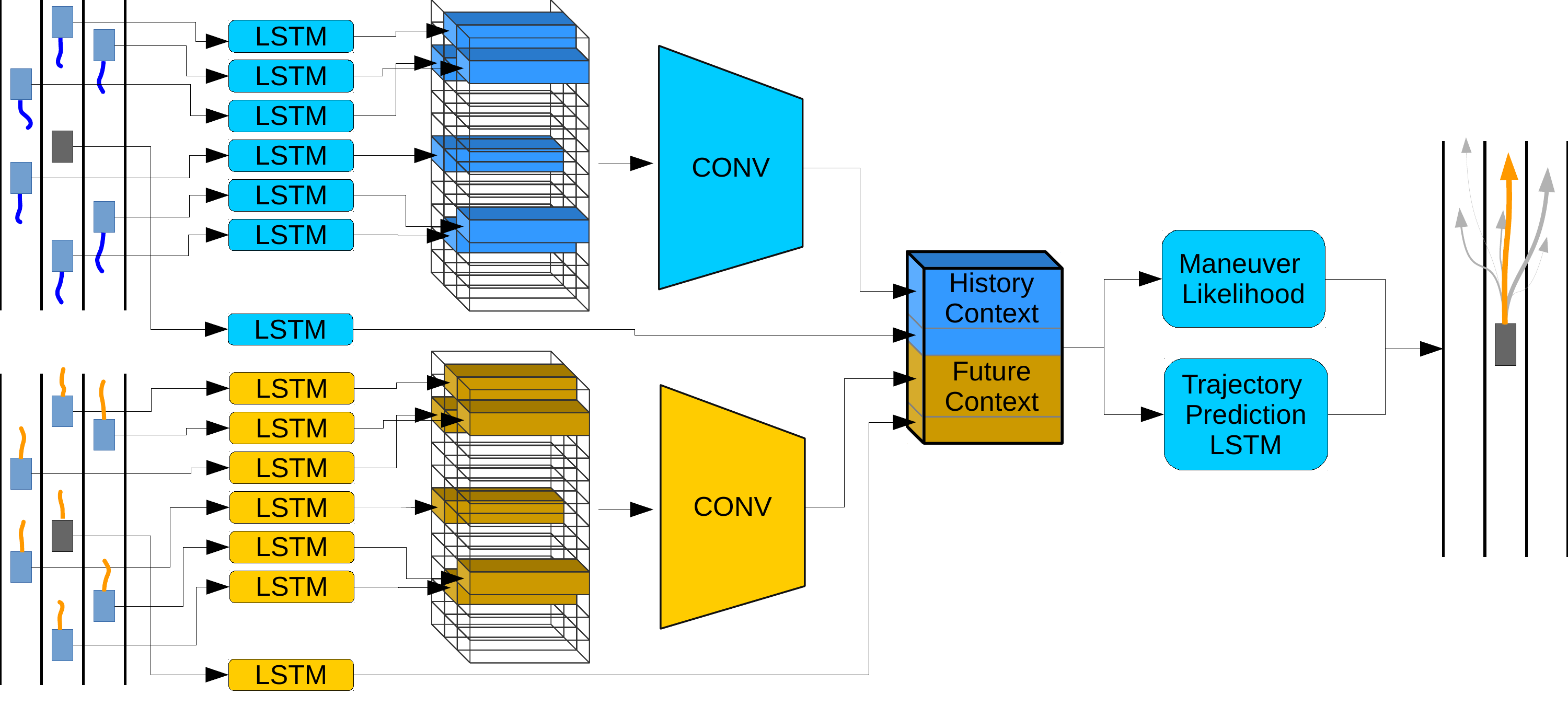}
      \caption{\textbf{Future-Conditional Convolutional Social Pooling ($\FCCSP$) Architecture.} All components shown in blue are part of the original $\CSP$ architecture as proposed by Deo and Trivedi \citep{Deo}. The yellow elements represent the portion of the model that processes the predicted future trajectories.}
    \label{fig:FC-CSP_architecture}
    \end{figure}

The \emph{Convolutional Social Pooling ($\CSP$) model} was originally proposed by Deo and Trivedi \citep{Deo}.
$\CSP$ combines a Long Short-Term Memory network (LSTM) encoder-decoder architecture with a convolution neural network (CNN) ``social pooling'' architecture to generate multimodal trajectory predictions.
    This model is depicted in blue in Figure~\ref{fig:FC-CSP_architecture}. For each target, $\CSP$ accepts as input the state-histories of both the target vehicle and its neighbors (vehicles that fall in a rectangular region around the target). 
    The final output is a set of six Gaussian distributions, each with an associated likelihood, that represent six possible trajectories.
        In our experiments, we take only the mode with the highest probability for prediction.
        In contrast to $\CV$, $\CSP$ is a ``high-fidelity'' motion prediction model.

The \emph{Future-Conditional Convolutional Social Pooling ($\FCCSP$) model}, shown in Figure \ref{fig:FC-CSP_architecture}, is a novel extension of $\CSP$, allowing the model to explicitly condition on predicted motion in addition to the motion history.
        This is accomplished by adding a ``future'' social pooling block (lower block), which is architecturally identical to the history social pooling block but receives as input the predicted future trajectories of the vehicles surrounding the target vehicle.
     The decoder LSTM layer in $\FCCSP$ receives the concatenation of both past and future ``social context'' tensors, and outputs a multimodal Gaussian trajectory distribution of the same form as the output of $\CSP$.
        As with $\CSP$, our experiments use only the highest probability mode.

%%%%%%%%%%%%%%%%%%%%%%%%%%%%%%%%%%%%%%

\section{Experiments}\label{sec:Experiments}
We perform three experiments to evaluate the performance of Multi-Fidelity Recursive Behavior Prediction.
    The first experiment is designed to compare our model against the existing state-of-the-art (the $\CSP$ baseline), while the other two experiments are intended to evaluate our algorithm in a setting that is more representative of an autonomous driving scenario.

All three experiments are conducted on the publicly available NGSIM I-80~\citep{I80} and US101~\citep{US101} datasets.
    There are 3 subsets of both US-101 and I-80, consisting of vehicle trajectories recorded by overhead camera %and processed to extract the 2D $(x,y)$ coordinates of each vehicle 
    at a frequency of 10 Hz.
    The test set consists of one quarter of the vehicle trajectories from each of the subsets of the US-101 and I-80 datasets. 
        We split the trajectories into segments of 8 s (sampled at 10 Hz), where we use 3 s of track history and a 5 s prediction horizon. 
We use the common metric of \emph{Root Mean Square Error} (RMSE) for evaluating the performance of our models.

\subsection{Experiment 1: Level 1 Recursive Behavior Prediction} % (\modelA)}
Our first experiment compares a specific implementation of Multi-Fidelity Recursive Behavior Prediction against the performance of the $\CSP$ baseline.
    This implementation is called Level $1$ Recursive Behavior Prediction (\modelA).  
    In \modelA, all agents are assigned a reasoning level of $1$.
    The level $0$ policy model for each agent is \emph{Convolutional Social Pooling} ($\CSP$), and the level $1$ policy for each agent is \emph{Future-Conditional Convolutional Social Pooling} ($\FCCSP$).
    Formally: ($\level_i, \policy_{i,0}, \policy_{i,1}) := (1, \CSP, \FCCSP) \ \forall i \in \lbrace 1, \ldots, n \rbrace$.

We train both the $\CSP$ (level $0$ policy) model and $\FCCSP$ (level $1$ policy) models jointly from scratch.
    As in the original $\CSP$ implementation, we use the leaky-ReLU activation with $\alpha=0.1$ for all layers and use Adam for optimization.

\vspace{-2mm}
\paragraph{Results for Experiment 1}
We compare the output of (\modelA) against the $\CSP$ baseline.
 The left part of Table~\ref{table:RMSE} %and Table~\ref{table:NLL} 
 shows RMSE values obtained at varying time horizons for $\CSP$ and \modelA. Note that $\CSP$$^\dotplus$ corresponds to the original values reported by Deo and Trivedi \citep{Deo} and $\CSP$$^\ast$ is our own implementation of the baseline $\CSP$ model.

\begin{table*}[t!]
\centering
    \caption{
    \textbf{Comparison on RMSE.}}
    \vspace{-6pt}
    \scalebox{0.9}{
    \begin{tabular}{ccccc}
       Prediction & \textbf{Baselines} & \textbf{Experiment 1} & \textbf{Experiment 2} & \textbf{Experiment 3}\\
      \begin{tabular}[t]{c}
          {Horizon (s)} \\
          \midrule
            1 \\
            2 \\
            3 \\
            4 \\
            5 \\
          \bottomrule
        %  \label{table:dummy} 
      \end{tabular}
       &
      \begin{tabular}[t]{cc}
          %\toprule
          {$\CSP$$^\dotplus$} & {$\CSP$$^*$} \\
          \midrule
           0.62 & 0.54  \\ 
           1.29 & 1.20  \\ 
           2.13 & 2.03  \\ 
           3.20 & 3.09  \\ 
           4.52 & 4.39  \\ 
          \bottomrule
        %   \label{table:ego_agno_rmse}
      \end{tabular}
       &
      \begin{tabular}[t]{c}
          %\toprule
          {{\modelA}} \\
          \midrule
           \textbf{0.53} \\ 
           \textbf{1.19} \\ 
           \textbf{1.95} \\ 
           \textbf{2.87} \\ 
           \textbf{3.97} \\ 
          \bottomrule
        %   \label{table:ego_agno_rmse}
      \end{tabular}
         &
         \begin{tabular}[t]{c}
          %\toprule
          {{\modelB}}  \\
          \midrule
            0.54  \\
            1.20  \\
            1.99  \\
            2.97  \\
            4.16  \\
        \bottomrule
     \end{tabular}
        &
      \begin{tabular}[t]{c}
          %\toprule
          {{\modelB} (planning)}  \\
          \midrule
            \textit{0.54}  \\
            \textit{1.19}  \\
            \textit{1.95}  \\
            \textit{2.88}  \\
            \textit{4.01}  \\
          \bottomrule
        %   \label{table:range_rmse}
      \end{tabular}
    \end{tabular}
    \label{table:RMSE}
    }
\end{table*}

\subsection{Experiment 2: Level 1 Multi-Fidelity Recursive Behavior Prediction}
For experiment 2, we introduce \emph{Level 1 Multi-Fidelity Recursive Behavior Prediction (\modelB)} (\modelB).
    \modelB targets the ``ego-centric'' prediction task, which is more representative of the autonomous driving use case:
        We randomly select a vehicle to treat as an ``ego'' agent, limiting the set of other agents in the scene to those within a plausible ``sensor range'' of this agent.
            This is repeated for many different ``ego'' agents during both training and testing.
    \modelB is identical to \modelA, except that it incorporates a multi-fidelity scheme wherein agents at the periphery of the designated ego agent's sensor range are assigned to a lower reasoning level ($\level_i = 0$) and a lower fidelity ($\policy_{i,0} = \CV$) policy model.
The \modelB $\CSP$ and $\FCCSP$ policies are jointly trained. The training process includes generating and using lower-fidelity constant velocity trajectory predictions for the peripheral agents.

\vspace{-2mm}
\paragraph{Results for Experiment 2}
It would take too long to exhaustively evaluate \modelB on the full test set (i.e. by treating every single vehicle in turn as the ego agent).
    Instead, we sample enough ego agents to ensure that a single level $1$ prediction can be computed for each vehicle in the test set.
        We report the average results of the level $1$ predictions from $10$ full iterations through the test set in this manner.
Our results are presented in Table \ref{table:RMSE} %and \ref{table:NLL} 
alongside the results from Experiment 1. 
\modelB exhibits slight improvement over $\CSP$, indicating that a multi-fidelity recursive prediction scheme can enhance performance even if the ``low-fidelity'' models are very naive.
    Improvement is more pronounced over longer prediction horizons. 

\subsection{Experiment 3: \modelB conditioned on Ego Future}
Experiment 3 seeks to quantify the performance improvement that results from conditioning motion prediction on a candidate future ego trajectory.
    To explore this question, we use the ground truth future trajectory as a surrogate for the ego agent's ``planned'' trajectory.
        This takes the place of the ego agent's level $0$ trajectory in a ``planning-aware'' version of \modelB. 
        In other words, we ``cheat'' by allowing the model to observe the ground truth future trajectory for each designated ego agent during training and testing.
            We wish to make it clear, therefore, that experiment 3 is not meant to compete with the other models.

\vspace{-2mm}
\paragraph{Results for Experiment 3}
RMSEs for planning-aware \modelB are shown in italics in the last column of Table~\ref{table:RMSE}. The numbers are better than for \modelB, which suggests that conditioning on a planned trajectory (as in a real automated driving scenario) can improve motion prediction.

%%%%%%%%%%%%%%%%%%%%%%%%%%%%%%%%%%
\section{Conclusion}\label{sec:Conclusion}
We have demonstrated that motion prediction in traffic scenes can be improved by recursively reasoning about future interaction between agents.
    We have also shown that multi-fidelity modeling can be effectively incorporated in the recursive prediction process.

    Immediate directions for future work include extending our method to reason about multiple possible future scenarios (i.e. multi-modal scene motion prediction), incorporating a more flexible and diverse set of policy models, reasoning about input state uncertainty, and devising a more comprehensive set of experiments and performance metrics to evaluate our models.
        We aim to eventually implement a refined version of our algorithm on a real automated vehicle.

%%%%%%%%%%%%%%%%%%%%%%%%%%%%%%%%%%%%%%%%%%%5

\paragraph{Acknowledgment}
We appreciate the support of Prof. Mykel J. Kochenderfer for his thorough reviews and insightful suggestions.

%\section*{References}
{\small
\bibliographystyle{plainnat}
\bibliography{egbib}

\begin{thebibliography}{21}
\providecommand{\natexlab}[1]{#1}
\providecommand{\url}[1]{\texttt{#1}}
\expandafter\ifx\csname urlstyle\endcsname\relax
  \providecommand{\doi}[1]{doi: #1}\else
  \providecommand{\doi}{doi: \begingroup \urlstyle{rm}\Url}\fi

\bibitem[Altché and de~La~Fortelle(2017)]{Altche2018}
F.~Altché and A.~de~La~Fortelle.
\newblock An {LSTM} network for highway trajectory prediction.
\newblock In \emph{IEEE International Conference on Intelligent Transportation
  Systems (ITSC)}, pages 353--359, 2017.
\newblock \doi{10.1109/ITSC.2017.8317913}.

\bibitem[Ammoun and Nashashibi(2009)]{Ammoun2009}
S.~Ammoun and F.~Nashashibi.
\newblock Real time trajectory prediction for collision risk estimation between
  vehicles.
\newblock In \emph{IEEE International Conference on Intelligent Computer
  Communication and Processing}, pages 417--422, 2009.
\newblock \doi{10.1109/ICCP.2009.5284727}.

\bibitem[Barth and Franke(2008)]{Barth2008}
A.~Barth and U.~Franke.
\newblock Where will the oncoming vehicle be the next second?
\newblock In \emph{IEEE Intelligent Vehicles Symposium}, pages 1068--1073,
  2008.
\newblock \doi{10.1109/IVS.2008.4621210}.

\bibitem[Camerer et~al.(2004)Camerer, Ho, and Chong]{Camerer2004}
Colin~F. Camerer, Teck-Hua Ho, and Juin-Kuan Chong.
\newblock A cognitive hierarchy model of games.
\newblock \emph{The Quarterly Journal of Economics}, 119\penalty0 (3):\penalty0
  861--898, 2004.
\newblock \doi{10.1162/0033553041502225}.
\newblock URL \url{http://dx.doi.org/10.1162/0033553041502225}.

\bibitem[Colyar and Halkias(2007{\natexlab{a}})]{I80}
J.~Colyar and J.~Halkias.
\newblock {US} highway {I}-80 dataset.
\newblock \emph{Federal Highway Administration (FHWA), Tech. Rep.}, \penalty0
  (FHWA-HRT07-030), 2007{\natexlab{a}}.

\bibitem[Colyar and Halkias(2007{\natexlab{b}})]{US101}
J.~Colyar and J.~Halkias.
\newblock {US} highway 101 dataset.
\newblock \emph{Federal Highway Administration (FHWA), Tech. Rep.}, \penalty0
  (FHWA-HRT07-030), 2007{\natexlab{b}}.

\bibitem[Costa-Gomes et~al.(2001)Costa-Gomes, Crawford, and
  Broseta]{Costa-Gomes2001}
Miguel Costa-Gomes, Vincent~P Crawford, and Bruno Broseta.
\newblock {Cognition and Behavior in Normal-form Games: An Experimental Study}.
\newblock \emph{Econometrica}, 69\penalty0 (5):\penalty0 1193--1235, 2001.
\newblock URL \url{http://econweb.ucsd.edu/{~}vcrawfor/CGCrBr01EMT.pdf}.

\bibitem[Deo and Trivedi(2018{\natexlab{a}})]{Deo}
Nachiket Deo and Mohan~M Trivedi.
\newblock {Convolutional Social Pooling for Vehicle Trajectory Prediction}.
\newblock pages 1468--1476, 2018{\natexlab{a}}.
\newblock URL \url{https://arxiv.org/pdf/1805.06771.pdf}.

\bibitem[Deo and Trivedi(2018{\natexlab{b}})]{Deo2018MultiModalTP}
Nachiket Deo and Mohan~Manubhai Trivedi.
\newblock Multi-modal trajectory prediction of surrounding vehicles with
  maneuver based lstms.
\newblock \emph{IEEE Intelligent Vehicles Symposium (IV)}, pages 1179--1184,
  2018{\natexlab{b}}.

\bibitem[Galceran et~al.(2017)Galceran, Cunningham, Eustice, and
  Olson]{Galceran2017}
Enric Galceran, Alexander~G. Cunningham, Ryan~M. Eustice, and Edwin Olson.
\newblock {Multipolicy decision-making for autonomous driving via
  changepoint-based behavior prediction: Theory and experiment}.
\newblock \emph{Autonomous Robots}, 41\penalty0 (6):\penalty0 1367--1382, 2017.
\newblock ISSN 0929-5593.
\newblock \doi{10.1007/s10514-017-9619-z}.
\newblock URL \url{http://link.springer.com/10.1007/s10514-017-9619-z}.

\bibitem[Hillenbrand et~al.(2006)Hillenbrand, Spieker, and
  Kroschel]{Hillenbrand2006}
Jrg Hillenbrand, Andreas~M. Spieker, and Kristian Kroschel.
\newblock {A Multilevel Collision Mitigation Approach--Its Situation
  Assessment, Decision Making, and Performance Tradeoffs}.
\newblock \emph{IEEE Transactions on Intelligent Transportation Systems},
  7\penalty0 (4):\penalty0 528--540, 2006.
\newblock ISSN 1524-9050.
\newblock \doi{10.1109/TITS.2006.883115}.
\newblock URL \url{http://ieeexplore.ieee.org/document/4019437/}.

\bibitem[{Hong Yoo} and Langari(2012)]{Yoo2012}
Je~{Hong Yoo} and Reza Langari.
\newblock {Stackelberg Game Based Model of Highway Driving}.
\newblock In \emph{ASME Annual Dynamic Systems and Control Conference}, pages
  499--508. ASME, 2012.
\newblock ISBN 978-0-7918-4529-5.
\newblock \doi{10.1115/DSCC2012-MOVIC2012-8703}.
\newblock URL
  \url{http://proceedings.asmedigitalcollection.asme.org/proceeding.aspx?doi=10.1115/DSCC2012-MOVIC2012-8703}.

\bibitem[{Hong Yoo} and Langari(2013)]{HongYoo}
Je~{Hong Yoo} and Reza Langari.
\newblock {A Stackelberg Game Theoretic Driver Model for Merging}.
\newblock In \emph{ASME Dynamic Systems and Control Conference}, 2013.
\newblock \doi{10.0/Linux-x86_64}.

\bibitem[Kaempchen et~al.(2004)Kaempchen, Wei\ss, Schaefer, and
  Dietmayer]{Kaempchen2004IMMOT}
Nico Kaempchen, Kilian Wei\ss, Marvin Schaefer, and K.~Dietmayer.
\newblock {IMM} object tracking for high dynamic driving maneuvers.
\newblock \emph{IEEE Intelligent Vehicles Symposium}, pages 825--830, 2004.

\bibitem[Khosroshahi et~al.(2016)Khosroshahi, Ohn-Bar, and
  Trivedi]{Khosroshahi}
A.~Khosroshahi, E.~Ohn-Bar, and M.~M. Trivedi.
\newblock Surround vehicles trajectory analysis with recurrent neural networks.
\newblock In \emph{IEEE International Conference on Intelligent Transportation
  Systems (ITSC)}, pages 2267--2272, Nov 2016.
\newblock \doi{10.1109/ITSC.2016.7795922}.

\bibitem[Miller and {Qingfeng Huang}(2002)]{Miller}
R.~Miller and {Qingfeng Huang}.
\newblock {An adaptive peer-to-peer collision warning system}.
\newblock In \emph{IEEE Vehicular Technology Conference}, volume~1, pages
  317--321, 2002.
\newblock ISBN 0-7803-7484-3.
\newblock \doi{10.1109/VTC.2002.1002718}.
\newblock URL \url{http://ieeexplore.ieee.org/document/1002718/}.

\bibitem[Morton et~al.(2017)Morton, Wheeler, and Kochenderfer]{Morton2017}
Jeremy Morton, Tim~A. Wheeler, and Mykel~J. Kochenderfer.
\newblock {Analysis of Recurrent Neural Networks for Probabilistic Modeling of
  Driver Behavior}.
\newblock \emph{IEEE Transactions on Intelligent Transportation Systems},
  18\penalty0 (5):\penalty0 1289--1298, 2017.
\newblock ISSN 1524-9050.
\newblock \doi{10.1109/TITS.2016.2603007}.
\newblock URL \url{http://ieeexplore.ieee.org/document/7565491/}.

\bibitem[Phillips et~al.(2017)Phillips, Wheeler, and
  Kochenderfer]{Phillips2017}
Derek~J. Phillips, Tim~A. Wheeler, and Mykel~J. Kochenderfer.
\newblock {Generalizable intention prediction of human drivers at
  intersections}.
\newblock In \emph{IEEE Intelligent Vehicles Symposium}, 2017.
\newblock ISBN 9781509048045.
\newblock \doi{10.1109/IVS.2017.7995948}.

\bibitem[Polychronopoulos et~al.(2007)Polychronopoulos, Tsogas, Amditis, and
  Andreone]{Polychronopoulos2007}
A.~Polychronopoulos, M.~Tsogas, A.J. Amditis, and L.~Andreone.
\newblock {Sensor Fusion for Predicting Vehicles' Path for Collision Avoidance
  Systems}.
\newblock \emph{IEEE Transactions on Intelligent Transportation Systems},
  8\penalty0 (3):\penalty0 549--562, 2007.
\newblock ISSN 1524-9050.
\newblock \doi{10.1109/TITS.2007.903439}.
\newblock URL \url{http://ieeexplore.ieee.org/document/4298909/}.

\bibitem[Sadigh et~al.(2016)Sadigh, Sastry, Seshia, and Dragan]{Sadigh2016}
Dorsa Sadigh, Shankar Sastry, Sanjit~A Seshia, and Anca~D Dragan.
\newblock {Planning for Autonomous Cars that Leverage Effects on Human
  Actions}.
\newblock \emph{Proceedings of Robotics: Science and Systems}, 2016.
\newblock \doi{10.15607/RSS.2016.XII.029}.
\newblock URL
  \url{https://pdfs.semanticscholar.org/e3e1/a23e0096f8ca51b783812310d44673a3c89a.pdf}.

\bibitem[Schmerling et~al.(2018)Schmerling, Leung, Vollprecht, and
  Pavone]{Schmerling2017}
Edward Schmerling, Karen Leung, Wolf Vollprecht, and Marco Pavone.
\newblock {Multimodal Probabilistic Model-Based Planning for Human-Robot
  Interaction}.
\newblock In \emph{IEEE International Conference on Robotics and Automation},
  2018.
\newblock URL \url{http://arxiv.org/abs/1710.09483}.

\end{thebibliography}
}

\end{document}